\documentclass[runningheads]{llncs}
\usepackage{graphicx}
\usepackage{subfigure}
\usepackage{bm}
\usepackage{amsmath}
\usepackage{amssymb}
\usepackage{graphicx}
\usepackage{color}
\usepackage{url}
\usepackage{multirow}
\usepackage{booktabs}
\usepackage{multirow}
\usepackage{wrapfig}
\usepackage{caption}
\usepackage{cite}
\usepackage{bbding}
\begin{document}
\title{Evaluation of Retinal Image Quality Assessment Networks in Different Color-spaces}
%
%
\author{Huazhu~Fu\inst{1} \and
Boyang~Wang\inst{1} \and
Jianbing~Shen\inst{1}\thanks{Corresponding author. (jianbing.shen@inceptioniai.org)} \and
Shanshan~Cui\inst{1} \and
Yanwu~Xu\inst{2} \and
Jiang~Liu\inst{3,4} \and
Ling~Shao\inst{1}}

 
\authorrunning{Fu et al.}

\institute{Inception Institute of Artificial Intelligence, Abu Dhabi, UAE \and
Baidu Inc., Beijing, China \and
Southern University of Science and Technology, Shenzhen, China \and
Cixi Institute of Biomedical Engineering, CAS, Ningbo, China  \\
Project page: \textcolor{red}{\url{https://github.com/hzfu/EyeQ}}
} 

\maketitle              
\begin{abstract}

Retinal image quality assessment (RIQA) is essential for controlling the quality of retinal imaging and guaranteeing the reliability of diagnoses by ophthalmologists or automated analysis systems. Existing RIQA methods focus on the RGB color-space and are developed based on small datasets with binary quality labels (i.e., `Accept' and `Reject'). In this paper, we first re-annotate an Eye-Quality (EyeQ) dataset with 28,792 retinal images from the EyePACS dataset, based on a three-level quality grading system (i.e., `Good', `Usable' and `Reject') for evaluating RIQA methods. Our RIQA dataset is characterized by its large-scale size, multi-level grading, and multi-modality. Then, we analyze the influences on RIQA of different color-spaces, and propose a simple yet efficient deep network, named Multiple Color-space Fusion Network (MCF-Net), which integrates the different color-space representations at both a feature-level and prediction-level to predict image quality grades. Experiments on our EyeQ dataset show that our MCF-Net obtains a state-of-the-art performance, outperforming the other deep learning methods. Furthermore, we also evaluate diabetic retinopathy (DR) detection methods on images of different quality, and demonstrate that the performances of automated diagnostic systems are highly dependent on image quality.

\keywords{Retinal image  \and Quality assessment \and Deep learning.}
\end{abstract}
\section{Introduction}

Retinal images are  widely used for early screening and diagnosis of several eye diseases, including diabetic retinopathy (DR), glaucoma, and age-related macular degeneration (AMD). However, retinal images captured using different cameras, by people with various levels of experience, have a large variation in  quality. A study based on UK BioBank showed that more than 25\% of the retinal images are not of high enough quality to allow accurate diagnosis~\cite{MacGillivray2015}. The quality degradation of retinal images, e.g., from inadequate illumination, noticeable blur and low contrast, may prevent a reliable medical diagnosis by ophthalmologists or automated analysis systems~\cite{Cheng2018TMI}. Thus, retinal image quality assessment (RIQA)  is required for controlling the quality of retinal image. However, RIQA is a subjective task that depends on the experience of the ophthalmologists and the type of eye disease. Moreover, the traditional general quality assessment methods for natural images are not suitable for the RIQA task.

Recently, several methods for RIQA specifically have been proposed, which can be divided into two main categories: structure-based methods and feature-based methods. \textbf{Structure-based methods} employ segmented structures to determine the quality of retinal images. For example, an image structure clustering method was proposed to extract compact representations of retinal structures to determine image quality levels~\cite{Niemeijer2006}.  Blood vessel structures are also widely used for identifying the quality of retinal images~\cite{Tobin2009,kohler2013,MacGillivray2015}. However,  structure-based methods rely heavily on the performance of structure segmentation, and cannot obtain latent visual features from images. \textbf{Feature-based methods}, on the other hand, directly extract feature representations from images, without structure segmentation. 
For example, features quantifying image color, focus, contrast and illumination can be calculated to represent the quality grade~\cite{PiresDias2014}. 
Wang~\textit{et.al.}~employed features based on the human visual system, with a support vector machine (SVM) or a decision tree to identify high-quality images~\cite{Wang2016}. 
A fundus image quality classifier that analyzes illumination, naturalness, and structure was also provided to assess quality~\cite{Shao2017}. 
Recently, deep learning techniques that integrate multi-level representations have been shown to obtain significant performances in a wide variety of medical imaging tasks. A combination of unsupervised features from saliency maps and supervised deep features from convolutional neural networks (CNNs) have been utilized to predict the quality level of retinal images~\cite{Yu2017}. For instance, Zago~\textit{et.al.}~adapted a deep neural network by using the pre-trained model from ImageNet to deal with the quality assessment task~\cite{Zago2018}. 
Although these deep methods have successfully overcome the limitations of hand-crafted features, they nevertheless have several of their own drawbacks. First, they focus on the RGB color-space, without considering other color-spaces that from part of the human visual system. Second, the existing  RIQA datasets only contain binary labels (i.e., `Accept' and `Reject'), which is a coarse grading standard for complex clinical diagnosis. Third, the RIQA community lacks a large-scale dataset, which limits the development of RIQA related methods, especially for deep learning techniques, since these require large amounts of training data.

To address the above issues, in this paper, we discuss the influences on RIQA of different color-spaces in deep networks. We first re-annotate an \textbf{Eye-Quality (EyeQ) dataset} with 28,792 retinal images selected from the EyePACS dataset, using a three-level quality grading system (i.e., `Good', `Usable' and `Reject'). Our EyeQ dataset considers the differences between ophthalmologists and automated systems, and can be used to evaluate other related works, including quality assessment methods, the influence of image quality on disease diagnosis, and retinal image enhancement. Second, we analyze the influences on RIQA of different color-spaces, and propose a general \textbf{Multiple Color-space Fusion Network (MCF-Net)} for retinal image quality classification. Our MCF-Net utilizes multiple base networks to jointly learn image representations from different color-spaces and fuses the outputs of all the base networks, at both a feature-level and prediction-level, to produce the final quality grade. Experiments demonstrate that our MCF-Net outperforms the other deep learning methods. In addition, we also apply the EyeQ dataset to evaluate the performances of DR detection methods for images of various qualities.

\section{Eye-Quality Dataset}

\begin{figure}[!t]
	\center
	\includegraphics[width =1\textwidth ]{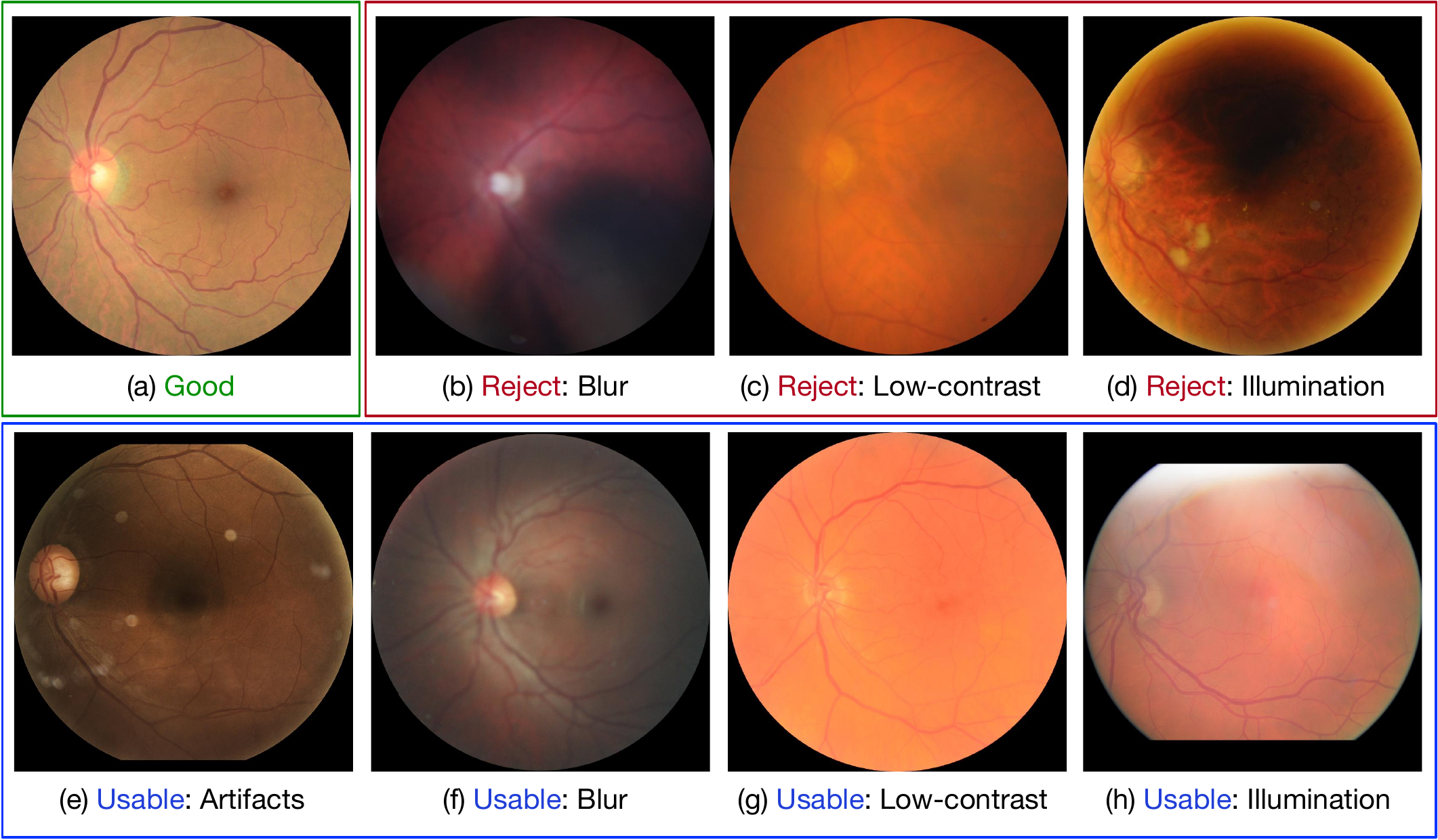} 
	\caption{Examples of different retinal image quality grades. The images  of `Good' quality (a)  provide clear diagnostic information, while the images of `Reject' quality (b-d) are insufficient for reliable diagnosis. However, some images are between `Good' and `Reject' (e-h), having some poor-quality indicators, but the main structures (e.g., disc and macular regions)  and lesion  are clear enough to be identified by ophthalmologists.}
	\label{img-quality}
\end{figure}

There are several publicly available RIQA datasets with manual quality annotations, such as HRF~\cite{kohler2013}, DRIMDB~\cite{Sevik2014}, and DR2~\cite{Pires2012}. However, they have various drawbacks. First, the image quality assessment of these datasets is based on binary labels, i.e., `Accept' and `Reject'. However, several images fall somewhere between these two categories.  For example, some retinal images of poor-quality, e.g., containing a few artifacts (Fig.~\ref{img-quality} (e)), or slightly blurred (Fig.~\ref{img-quality} (f)), are still gradable by clinicians, so should not be labeled as `Reject', but they may mislead automated medical analysis methods, so can also not be labeled as `Accept'. Second, retinal images of the existing RIQA datasets are often captured by the same camera, which can not be used to evaluate the robustness of RIQA methods against various imaging modalities. Third, the existing datasets are limited in size, and there lacks a large-scale quality grade dataset for developing deep learning methods. 

\begin{table*}[!t]
	\renewcommand{\arraystretch}{1.2}
	\centering
	\caption{Summary of our EyeQ dataset, where DR-i denotes the DR presence on grade i based on the labels in EyePACS dataset.}
	\begin{tabular}{|l||c|c|c|c|c|c||c|c|c|c|c|c|}
		\hline
		       &    \multicolumn{6}{c||}{ Training set }     &      \multicolumn{6}{c|}{ Testing set }   \\ \cline{2-13}
		       & DR-0  & DR-1 & DR-2  & DR-3 & DR-4 &  All   &  DR-0  & DR-1  & DR-2  & DR-3 & DR-4 &  All   \\ \hline
		Good   & 6,342 & 699  & 1,100 & 167  &  39  & 8,347  & 5,966  &  886  & 1,354 & 199  &  65  & 8,470  \\
		Usable & 1,353 & 103  &  283  &  79  &  58  & 1,876  & 3,201  &  359  &  721  & 145  & 133  & 4,559  \\
		Reject & 1,544 & 109  &  426  &  87  & 154  & 2,320  & 2,195  &  153  &  569  & 104  & 199  & 3,220  \\ \hline
		Total  & 9,239 & 911  & 1,809 & 333  & 251  & \textbf{12,543} & 11,362 & 1,398 & 2,644 & 448  & 397  & \textbf{16,249} \\ \hline
	\end{tabular}%
	\label{Tab_grade_QA}%
\end{table*}%

To address the above issues, in this paper, we re-annotate an Eye-Quality (EyeQ) dataset from the EyePACS dataset, which is a large retinal image dataset captured by different models and types of cameras, under a variety of imaging conditions. Our EyeQ dataset utilizes a three-level quality grading system by considering four common quality indicators, including blurring, uneven illumination, low-contrast, and artifacts. Our three quality grades are defined as: 
\begin{itemize}
	\item \textbf{`Good' grade}: the retinal image has no low-quality factors, and all retinopathy characteristics are clearly visible, as shown in Fig.~\ref{img-quality} (a).
	\item \textbf{`Usable' grade}: the retinal image has some slight low-quality indicators, which can not observe the whole image clearly (e.g., low-contract and blur) or affect the automated medical analysis methods (e.g., artifacts), but the main structures (e.g., disc, macula regions)  and lesion  are clear enough to be identified by ophthalmologists, as shown in Fig.~\ref{img-quality} (e-h). For the uneven illumination case, the readable region of fundus image is larger than 80\%.
	\item \textbf{`Reject' grade}: the retinal image has a serious quality issue and cannot be used to provide a full and reliable diagnosis, even by ophthalmologists, as shown in Fig.~\ref{img-quality} (b-d). Moreover, the fundus image with invisible disc or macula region is also be treated as `Reject' grade. 
\end{itemize}
To re-annotate the EyeQ dataset, we asked two experts to grade the quality of images in EyePACS. Then, the images with ambiguous labels were discarded, yielding a collection of 28,792 retinal images. A summary of this EyeQ dataset is given in Table~\ref{Tab_grade_QA}. Note that some images with `Reject' grades still have DR grade labels from the EyePACS dataset. The reason is that our quality standard is based on the diagnosability for general eye diseases, such as glaucoma, AMD, etc., rather than only DR. Although some low-quality images have visible lesions for DR diagnosis, they are not of high enough quality for diagnosing other diseases, lacking, for example, clear views of optic disc and cup regions for glaucoma screening, or visible macula regions for AMD analysis.

\section{Multiple Color-space Fusion Network}

\begin{figure}[!t]
	\center
	\includegraphics[width =1\textwidth ]{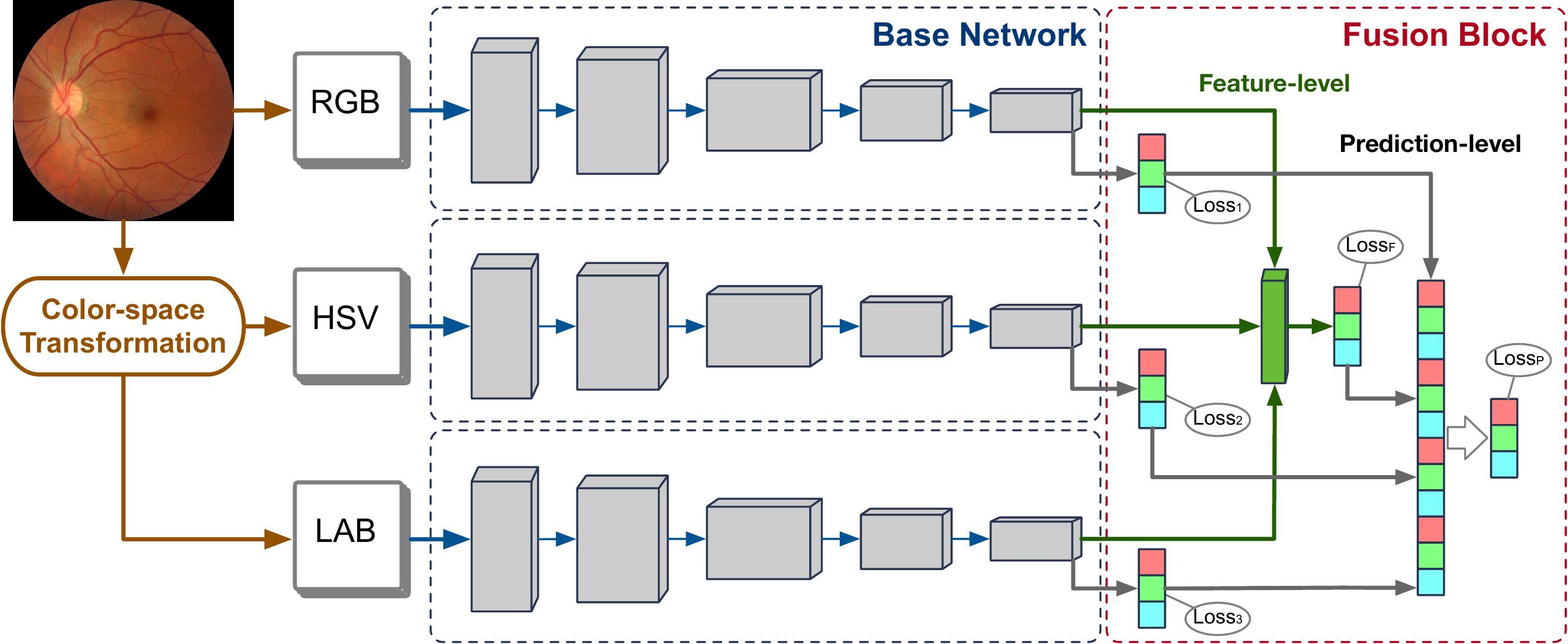} 
	\caption{The architecture of our MCF-Net, which contains multiple base networks for different color-spaces. A fusion block is employed to integrate the multiple outputs of these base networks on both feature-level and prediction-level.}
	\label{img-framework}
\end{figure}

Recently, deep learning techniques have been shown to obtain satisfactory performances in retinal image quality assessment~\cite{Yu2017,Zago2018}. However, these methods only focus on the RGB color-space, and ignore other color-spaces that included in the human visual system. A color-space identifies a particular combination of the color model and the mapping function. Different color-spaces represent different characteristics, and can be used to extract diverse visual features, which have been demonstrated to affect the performances of deep learning networks~\cite{MISHKIN201711}. 
In this paper, we analyze the influences of different color-spaces on the RIQA task, and propose a general Multiple Color-space Fusion Network (MCF-Net) to integrate the representations of various color-spaces. Besides the original RGB color-space, we also consider HSV and LAB color-spaces, which are widely used in computer vision tasks and are obtained through nonlinear conversions from the RGB color-space.

Fig.~\ref{img-framework} illustrates the architecture of our MCF-Net. The original RGB image is first transferred to HSV and LAB color-spaces, and fed into the base networks. The base networks generate image features by employing multi-scale CNN layers. Then, a fusion block is used to combine the output of each base network at both a feature-level and prediction-level. 
First, the feature maps from the base networks are concatenated and input to a fully connected layer to generate a feature-level fusion prediction. Then,  the predictions of all the base networks and feature-level fusion are concatenated and fed into a fully connected layer to produce the final prediction-level fusion result. Our two-level fusion block guarantees the full integration of the different color-spaces. On the other hand, our fusion block also maintains the independence and integrity of the base networks, which enables any deep network to be implemented as the base network. 
Different from existing deep fusion networks, which only use the loss function of the last layer to train the whole model, our MCF-Net retains all loss functions of the base networks, to improve their transparency for each color-space, and combines them with the fusion loss of the last layer, as:
\begin{equation}
	Loss_{total}=\sum_{i=1}^3 w_i Loss_{i} + w_F  Loss_{F} + w_P  Loss_{P},
\end{equation}
where $Loss_{i}$,  $Loss_{F}$ and $Loss_{P}$ denote the \textit{multi-class cross-entropy} loss functions of the base networks and the two-level fusion layers, respectively. $w_i$, $w_F$ and $w_m$ are trade-off weights, which are set to $w_i = 0.1$, $w_F= 0.1$ and $w_m = 0.6$ to highlight the final prediction-level fusion layer. 

\section{Experiments}

\noindent \textbf{Implementation Details:} For each input image, we first detect the retinal mask using the Hough Circle Transform, and then crop the mask region to reduce the influence of  black background. Finally, the image is resized to $224\!\times\!224$ and normalized to $[-1, 1]$, before being fed to our MCF-Net. For data augmentation, we apply vertical and horizontal flipping, random drifting and rotation.
The initial weights of the base networks are loaded from pre-trained models based on ImageNet, and the parameters of the final fully connected layer are randomly initialized. Our model is optimized using the SGD algorithm with a learning rate of 0.01. The framework is implemented on PyTorch.\\

\noindent \textbf{Experimental Settings:} Our EyeQ dataset is divided into a training set (12,543 images) and a testing set (16,249 images), following the EyePACS settings, as shown in Table~\ref{Tab_grade_QA}. We evaluate our MCF-Net  utilizing three state-of-the-art networks: ResNet18~\cite{resnet2016}, ResNet50~\cite{resnet2016}, and DenseNet121~\cite{densenet2017}. For each base network, we compare our MCF-Net in terms of the network with individual color-space (i.e., RGB, HSV and LAB). We also report the average result (AVG) when combining the predictions of the three color-spaces directly, without the fusion block. For the non-deep learning baseline, we implement the RIQA method from~\cite{Wang2016}, which is based on three visual characteristics (i.e., multi-channel sensation, just noticeable blur, and the contrast sensitivity function) and an SVM classifier with a radial based function.  For evaluation metrics, we employ average accuracy, precision, recall, and F-measure ($\frac{2 * precision * recall}{precision + recall}$).\\


\noindent \textbf{Results and Discussion:} The performances of different methods are reported in Table~\ref{Tab_result_QA}. We can make the following observations: (1) The performance of the non-deep learning baseline~\cite{Wang2016}  is obviously lower than those of deep learning based methods. This is reasonable because deep learning can extract highly discriminative representations from the retinal images directly, using multiple CNN layers, which are superior to the hand-crafted features in~\cite{Wang2016} and lead to better performance. (2) For the different color-spaces, the networks in RGB and LAB color-spaces perform better than that in HSV color-space. One possible reason is that the RGB color-space is closer to the raw data captured from the camera, and thus a more natural way to represent image data. Another possible reason is that the model pre-trained on ImageNet are based on RGB color-space, which is more suitable for fine-tuning in the same color-space. The LAB color-space represents the lightness and color components of green–red and blue–yellow. The lightness channel directly reflects the illustration status of images, which is the main quality indicator for retinal images. (3) Combinations of different color-spaces, even the simple average fusion (AVG), perform better than those of individual color-space. (4) Our MCF-Net outperforms the models with an individual color-spaces and average fusion. This demonstrates that the multi-level fusion block can produce a stable improvement to benefit the quality assessment task. Moreover, for deep learning models, DenseNet121-MCF obtains the best performance, outperforming ResNet18-MCF and ResNet50-MCF. \\

\begin{table*}[!t]
	\centering
	\caption{Performances of different methods on test set. }
	\begin{tabular}{|p{90pt}|c|c|c|c|}
		\hline
		& ~Accuracy~ & ~Precision~ & ~~~Recall~~~ & ~F-measure~ \\ \hline
		Baseline~\cite{Wang2016} &   0.8372   &   0.7404    &    0.6945    &   0.6991    \\ \hline
		ResNet18-RGB             &   0.8914   &   0.8044    &    0.8166    &   0.8087    \\
		ResNet18-HSV             &   0.8859   &   0.8010    &    0.7972    &   0.7980    \\
		ResNet18-LAB             &   0.8912   &   0.8071    &    0.8138    &   0.8083    \\
		ResNet18-AVG             &   0.8966   &   0.8164    &    0.8226    &   0.8176    \\
		ResNet18-MCS             &   0.9029   &   0.8457    &    0.8189    &   0.8288    \\ \hline
		ResNet50-RGB             &   0.8921   &   0.8123    &    0.8078    &   0.8100    \\
		ResNet50-HSV             &   0.8709   &   0.7706    &    0.7778    &   0.7735    \\
		ResNet50-LAB             &   0.8925   &   0.8078    &    0.8146    &   0.8091    \\
		ResNet50-AVG             &   0.8957   &   0.8156    &    0.8183    &   0.8163    \\
		ResNet50-MCS             &   0.9004   &   0.8389    &    0.8126    &   0.8230    \\ \hline
		DenseNet121-RGB          &   0.8943   &   0.8194    &    0.8114    &   0.8152    \\
		DenseNet121-HSV          &   0.8786   &   0.7963    &    0.7695    &   0.7808    \\
		DenseNet121-LAB          &   0.8882   &   0.8130    &    0.7937    &   0.8010    \\
		DenseNet121-AVG          &   0.8952   &   0.8240    &    0.8065    &   0.8143    \\
		DenseNet121-MCS          &   0.9175   &   0.8645    &    0.8497    &   0.8551    \\ \hline
	\end{tabular}%
	\label{Tab_result_QA}%
\end{table*}%

\noindent \textbf{DR Detection in Different-quality Images:} 
In this paper, we also apply our EyeQ dataset to evaluate the performances of DR detection methods for different-quality images. We train three deep learning networks, e.g., ResNet-18~\cite{resnet2016}, ResNet-50~\cite{resnet2016} and DenseNet-121~\cite{densenet2017}, on the whole  EyePACS training set, and evaluate their performances on images of different quality. Table~\ref{tab_dr} shows the accuracy scores of the DR detection models. As expected, the performances of the methods decrease along with the quality degradation of the images.  
The accuracy scores of ResNet-18~\cite{resnet2016}, ResNet-50~\cite{resnet2016} and DenseNet-121~\cite{densenet2017} decrease by 0.21\%, 0.04\%, and 0.23\%, respectively, from `Good' to `Usable', and by 1.45\%, 0.82\%, and 1.31\% from `Usable' to `Reject'. Note that since we train the DR detection methods on the whole EyePACS training set, which includes different quality images, the networks are somewhat robust to poor-quality images. However, poor-quality images still pose challenges for automated diagnosis systems, even for the images labeled as `Usable', which could provide diagnosable information to ophthalmologists.

\begin{table*}[!t]
	\centering
	\caption{Accuracy score of DR detection methods on different-quality images.}
	\begin{tabular}{|p{70pt}|c|c|c|}
		\hline & ~`Good'~ & ~`Usable'~ & ~`Reject'~ \\ \hline
		ResNet18                &  0.9014  &   0.8993   &   0.8848   \\
		ResNet50                &  0.9154  &   0.9150   &   0.9068   \\
		DenseNet121             &  0.9174  &   0.9151   &   0.9020   \\ \hline
	\end{tabular} 
	\label{tab_dr}%
\end{table*}%

\section{Conclusion}

In this paper, we have constructed an Eye-Quality (EyeQ) dataset from the EyePACS dataset, with a three-level quality grading system (i.e., `Good', `Usable' and `Reject'). Our EyeQ dataset has the advantages of a large-scale size, multi-level grading, and multi-modality. Moreover, we have also proposed a general Multiple Color-space Fusion Network (MCF-Net) for retinal image quality classification, which integrates different color-spaces. Experiments have demonstrated that MCF-Net outperforms other methods. In addition, we have also shown that image quality affects the performances of automated DR detection methods. 
We hope our work can draw more interest from the community to work on the RIQA task, which plays a critical role in applications such as retinal image segmentation~\cite{Fu2018_MNET,Gu2019}, and automated disease diagnosis~\cite{Fu2018_DENET}.

\bibliographystyle{splncs04}
\bibliography{paper0258}

\end{document}